\pdfoutput=1

\documentclass[11pt]{article}

\usepackage[final]{acl}

\usepackage{times}
\usepackage{latexsym}

\usepackage[T1]{fontenc}

\usepackage[utf8]{inputenc}

\usepackage{microtype}

\usepackage{inconsolata}

\usepackage{graphicx}

\usepackage{enumitem}
\setlist{nosep}
\usepackage{amsmath}
\usepackage{amssymb}
\usepackage{booktabs}
\usepackage{url}
\usepackage{multirow}
\usepackage{threeparttable} 
\usepackage[normalem]{ulem} 
\useunder{\uline}{\ul}{}
\usepackage{subcaption}    
\usepackage{makecell}
\usepackage{hyperref}

\usepackage{xspace}
\newcommand{\mname}{Hi-Merging\xspace}

\title{Training-free LLM Merging for Multi-task Learning}

\author{
  \textbf{Zichuan Fu\textsuperscript{1}\thanks{Co-first authors with equal contributions.}},
  \textbf{Xian Wu\textsuperscript{2}\footnotemark[1]},
  \textbf{Yejing Wang\textsuperscript{1}}, \\
  \textbf{Wanyu Wang\textsuperscript{1}},
  \textbf{Shanshan Ye\textsuperscript{3}},
  \textbf{Hongzhi Yin\textsuperscript{4}}, 
  \textbf{Yi Chang\textsuperscript{5}}, \\
  \textbf{Yefeng Zheng\textsuperscript{2,6}},
  \textbf{Xiangyu Zhao\textsuperscript{1}\thanks{Corresponding author.}}
\\
\\
  \textsuperscript{1} City University of Hong Kong
  \textsuperscript{2} Tencent Jarvis Lab  \\
  \textsuperscript{3} University of Technology Sydney
  \textsuperscript{4} University of Queensland  \\
  \textsuperscript{5} Jilin University
  \textsuperscript{6} Westlake University
\\
  \small{
  \texttt{
    \href{mailto:zc.fu@my.cityu.edu.hk}{zc.fu@my.cityu.edu.hk},
    \href{mailto:kevinxwu@tencent.com}{kevinxwu@tencent.com},
    \href{mailto:xianzhao@cityu.edu.hk}{xianzhao@cityu.edu.hk}
  }}
}

\begin{document}
\maketitle
\begin{abstract}
Large Language Models (LLMs) have demonstrated exceptional capabilities across diverse natural language processing (NLP) tasks.
The release of open-source LLMs like LLaMA and Qwen has triggered the development of numerous fine-tuned models tailored for various tasks and languages. In this paper, we explore an important question: is it possible to combine these specialized models to create a unified model with multi-task capabilities.
We introduces \textbf{H}ierarchical \textbf{I}terative \textbf{Merging} (Hi-Merging), a training-free method for unifying different specialized LLMs into a single model.
Specifically, Hi-Merging employs model-wise and layer-wise pruning and scaling, guided by contribution analysis, to mitigate parameter conflicts.
Extensive experiments on multiple-choice and question-answering tasks in both Chinese and English validate Hi-Merging's ability for multi-task learning. 
The results demonstrate that Hi-Merging consistently outperforms existing merging techniques and surpasses the performance of models fine-tuned on combined datasets in most scenarios. 
Code is available at \href{https://github.com/Applied-Machine-Learning-Lab/Hi-Merging}{Applied-Machine-Learning-Lab/Hi-Merging}.

\end{abstract}

\section{Introduction}
\label{introduction}

\begin{figure*}[ht]
    \centering
    \includegraphics[width=0.90\textwidth]{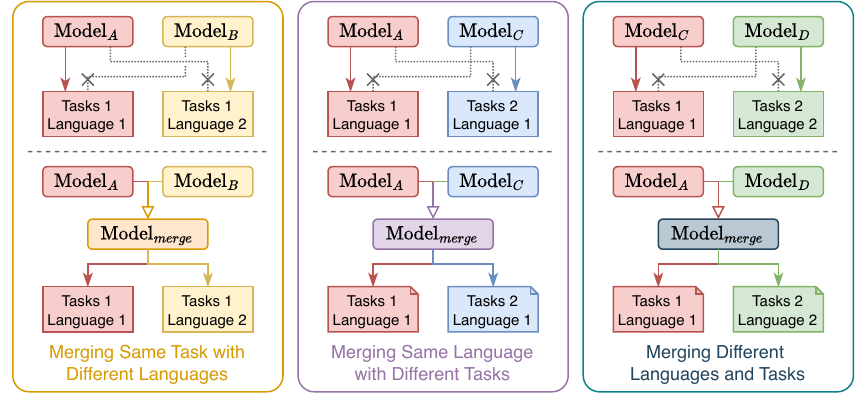}
    \caption{Illustration of three paradigms for our LLM merging: merging models that specialize in different languages (left), merging models that excel at different tasks (middle), and merging models that exhibit expertise in both different languages and different tasks (left). Through such merging, a single model can inherit the combined capabilities of both original models, enabling broader applicability and enhanced performance.}
    \label{fig:llm-merging}
\end{figure*}

Large Language Models (LLMs) have revolutionized Natural Language Processing (NLP) by demonstrating unprecedented capabilities in capturing and utilizing world knowledge \cite{llmsurvey}. Recent advances in architecture design and training methodologies have enabled models like GPT-4 \cite{gpt4} to engage in human-like dialogue and solve real-world problems, enabling breakthroughs in healthcare, recommender system, and scientific research~\cite{ESR,uni-ctr,hallu}.

With the advent of open-source large language models (LLMs) like LLaMA-3~\cite{llama3} and Qwen~\cite{qwen2}, significant research efforts have been dedicated to fine-tuning these models for specific tasks, domains, and languages~\cite{lsa}. As a result, Hugging Face\footnote{\href{https://huggingface.co/}{https://huggingface.co/}} now hosts over one million specialized LLMs across various tasks, and this number continues to grow rapidly. These models represent a vast repository of task-specific and language-specific expertise, ranging from medical applications~\cite{chen2023huatuogptii,MOELoRA} to financial question and answering~\cite{financial}. A natural question arises: is it possible to combine these task-specific fine-tuned LLMs into a single unified model with broad capabilities, including multi-lingual and multi-task functionalities? If achievable, the deployment of such a unified model could perform multiple tasks that currently require multiple LLMs, thereby significantly enhancing the application of LLMs.
One potential solution is to gather all fine-tuning data and retrain the LLMs from scratch. However, this approach has three significant disadvantages: 1) the availability of fine-tuning data, as the models are often public but the data is not~\cite{mtl}; 2) retraining large LLMs requires substantial computational resources; and 3) balancing the training data from different tasks to maintain strong performance across all tasks without compromising any individual task (avoiding the ``seesaw effect''~\cite{seesaw} where improving one task's performance leads to degradation in others) is a non-trivial challenge.

Based on the above considerations, model merging~\cite{modelmergingsurvey} emerges as a promising solution for unifying multiple specialized models while preserving their individual capabilities.
However, current model merging methods face two fundamental challenges. 
First, interference between merged models can arise from noise introduced by data bias~\cite{bias} or the training process, such as overfitting, impairs the merged model's generalization.
Second, models trained independently follow distinct optimization trajectories, leading to different knowledge alignments in their parameter spaces~\cite{taskarithmetic}. These misaligned parameters become incompatible for direct combination without additional training.

To address these challenges, we propose Hi-Merging, a \textbf{H}ierarchical \textbf{I}terative \textbf{Merging} method. It first applies model-wise pruning and scaling to the delta vectors (parameter differences between fine-tuned models and the foundation model) to eliminate noisy parameters introduced during fine-tuning.
Then, we apply layer-wise pruning and scaling iteratively for the knowledge misalignment, starting from the most conflicted layers. To identify the severity of layer-wise conflicts, we develop contribution analysis - a method that quantifies each layer's contribution by measuring how adding or removing specific layers affects model capabilities. By analyzing how our contribution metrics change before and after a pre-merging process, we can identify potential conflicts, thereby guiding our iterative optimization process to resolve parameter incompatibilities without additional training.

Our contributions can be summarized as follows:
\begin{itemize}[leftmargin=*]
    \item
    We investigate the challenges and potential of training-free model merging for integrating LLMs specialized in diverse tasks (e.g., MCQA, QA) and languages (e.g., English, Chinese), addressing a complex multi-task scenario.
    \item
    We propose Hi-Merging, a hierarchical iterative approach that effectively reduces the interference of noise and knowledge alignment conflicts during model merging.
    \item
    Extensive experiments on four datasets demonstrate the effectiveness of \mname in multi-task merging across different tasks and languages, consistently achieving superior performance.
\end{itemize}


\section{Preliminary for LLM Merging}
In this section, we detail notations and introduce existing LLM merging solutions as the preliminary. 

Model merging aims to combine multiple models with distinct capabilities as a single model, which has all the strengths of these models. In this paper, we user two-model merging for illustration: 
Given models $\mathcal{M}_A$ and $\mathcal{M}_B$ with parameters $\boldsymbol{\theta}_A$ and $\boldsymbol{\theta}_B$, both fine-tuned from a foundation model $\mathcal{M}_F$ with parameters $\boldsymbol{\theta}_F$ for tasks $t_A$ and $t_B$ respectively, model merging aims to combine them into a single model $\mathcal{M}_{\mathrm{merge}}$ with parameters $\boldsymbol{\theta}_{\mathrm{merge}}$ that preserves capabilities for both tasks.

Typical model merging strategies include weighted averaging and delta vector-based merging. The former combines model parameters through a weighted sum~\cite{modelsoup}: 
\begin{equation}
\label{eq:1}
    \boldsymbol{\theta}_{\mathrm{merge}}=\sum_{m\in \{A,B\}} \omega_m \boldsymbol{\theta}_m ,
\end{equation}
where $\omega_m$ is the weight to balance different capabilities constrained to $\sum_{m\in \{A,B\}} \omega_m=1, \omega_m>0$. And $m\in \{A,B\}$ is the model identifier.

The second strategy merges models based on delta vectors, the parameter differences between fine-tuned models and their foundation model, which can be mathematically defined as:
\begin{equation}
     \boldsymbol{\delta}_m = \boldsymbol{\theta}_m - \boldsymbol{\theta}_{F}\label{eq:delta_def}.
\end{equation}
Delta vectors $\boldsymbol{\delta}_m$ defined in Equation~\eqref{eq:delta_def} reveal model-specific updates from the foundation model, enabling a delta-weighted merging strategy~\cite{taskarithmetic}:
\begin{equation}
\label{eq:3}
\boldsymbol{\theta}_{\mathrm{merge}}=\boldsymbol{\theta}_{F} + \sum_{m\in \{A,B\}} \omega_m\boldsymbol{\delta}_m.
\end{equation}
where $\omega_m>0$. Note that both strategies, illustrated in Equation~\eqref{eq:1} and Equation~\eqref{eq:3}, can be easily extended to multiple model merging scenarios by expanding the model list $\{A, B\}$.

\section{Method}
\label{hierarchical-pruning-scaling}

\label{method}
In this section, we introduce the proposed method, which consists of two major components: (1) model-wise pruning and scaling that removes noisy and redundant parameters and moderate excessive ones and (2) layer-wise pruning and scaling iterating on conflicted layers to address knowledge misalignment issues.


\subsection{Model-wise Pruning and Scaling}
\label{model-wise-pruning-scaling}

This section introduces two operations to process delta vectors: pruning and scaling. 

During the fine-tuning, models can accumulate noisy parameters and learn sharp parameters for the specific fine-tuning task. We introduce the pruning and scaling operations to tackle these two problems, respectively, which are controlled by the following hyperparameters:

\begin{itemize}[leftmargin=*]
    \item \textbf{Pruning Threshold ($p$)}: This parameter specifies the proportion of the delta vector that should be preserved. 
    By retaining the largest $p$ percentage of the vector's components and rendering the remaining $(1 - p)$ to zero, the pruning operation can eliminates trivial parameter updates (data-specific noise) while preserving meaningful task-specific knowledge.

    
    \item \textbf{Scaling Factor ($s$)}: This factor controls the magnitude of the delta vector. With this parameter, the scaling operation contributes to addressing over-aggressive parameters by scaling down sharp updates, which may result from the overfitting during fine-tuning. The pruning does not apply to large parameter changes as they likely encode essential knowledge. The scaling provides a way to moderate their excessive influence. 
\end{itemize}
With these hyperparameters, the pruning and scaling cooperatively process the delta vectors in a complementary manner: pruning eliminates negligible parameter changes while scaling moderates the significant ones. Note that both $p$ and $s$ constrained to $[0, 1]$.

\begin{figure}[t]
    \centering
    \includegraphics[width=\columnwidth]{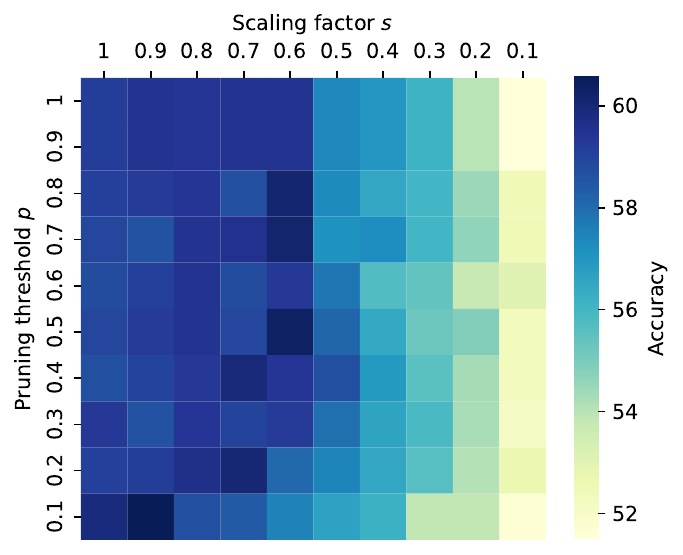}
    \caption{The accuracy of the fine-tuned Qwen2-7B-Instruct on the MedQA dataset after the model-wise pruning and scaling process with different combinations of the pruning threshold $p$ and the scaling factor $s$.}
    \label{fig:heat}
\end{figure}

We empirically validate the effectiveness of the pruning and scaling 
operations by iterating $p$ and $s$ from $[0.1,1]$. The result is visualized in Figure~\ref{fig:heat}. We can find that the individual model can maintain or even improve performance with appropriate pruning and scaling. For example, $p=0.1, s=0.9$ (preserving 10\% of parameters and scaling all delta values with 0.9) can defeat the original model ($p=1, s=1$). This finding supports our idea of conducting model-wise pruning and scaling to overcome noisy and radical parameter updates.


Next, we introduce the model-wise pruning and scaling details. Specifically, the delta vector (defined in Equation~\eqref{eq:delta_def}) for a given LLM $\mathcal{M}_m$ can be defined as $\boldsymbol{\delta}_m = [\delta_{m,1}, \delta_{m,2}, \ldots, \delta_{m,N}]$, where $m \in \{A, B\}$ is the model identifier and $N$ indicates the size of trainable parameters.

The \textbf{pruning} operation $\operatorname{Top}_p$ retains the $\lceil p \cdot N \rceil$ elements of $\boldsymbol{\delta}_m$ with the largest absolute value and zeros out the rest, resulting in $\tilde{\boldsymbol{\delta}}_m$:
\begin{equation}
    \tilde{\boldsymbol{\delta}}_m = \operatorname{Top}_p(\boldsymbol{\delta}_m).
\end{equation}
In detail, the $n$-th component of $\tilde{\boldsymbol{\delta}}_m$ is:
\begin{equation}
    \scalebox{0.85}{$
    \tilde{\delta}_{m, n}= \begin{cases}\delta_{m, n}, & \text { if } n \in\{\pi(1), \pi(2), \ldots, \pi(\lceil p \cdot N\rceil)\} \\ 0, & \text { otherwise }\end{cases}
    $},
\end{equation}
where $\pi(n)$ represents the index of the $n$-th largest component of $\boldsymbol{\delta}_m$ in absolute value, such that:
\begin{equation}
    \left|\delta_{m, \pi(1)}\right| \geq\left|\delta_{m, \pi(2)}\right| \geq \cdots \geq\left|\delta_{m, \pi(N)}\right|.
\end{equation}

The \textbf{scaling} operation adjusts the magnitude of the pruned delta vector $\tilde{\boldsymbol{\delta}}_m$ by multiplying it with the scaling factor $s \in[0,1]$ as $s\tilde{\boldsymbol{\delta}}_m$. 

Regarding the different setting of $p$ and $s$ for each model, the model-wise pruning and scaling can be compactly expressed as:
\begin{equation}
    \boldsymbol{\hat\delta}_m=s_m \cdot \operatorname{Top}_{p_{m}}\left(\boldsymbol{\delta}_m\right)=s_m\tilde{\boldsymbol{\delta}}_m,
\end{equation}
where $\boldsymbol{\hat\delta}_m$ represents the delta vector after the model-wise pruning and scaling.

Through model-wise process with pruning and scaling, we effectively identify noisy and excessive parameter updates from the fine-tuning, maintaining and moderating the key knowledge about the fine-tuning task for the subsequent merging.


\subsection{Layer-wise Pruning and Scaling}
\label{layer-wise-pruning-scaling}

In this section, we conduct the layer-wise model merging with pruning and scaling operations with a novel contribution analysis method to measure the parameter conflict. 

\subsubsection{Contribution Analysis}
Directly merging the model-wise processed delta vectors $\{\boldsymbol{\hat\delta}_m\}_{m\in\{A,B\}}$ as in Equation~\ref{eq:1} or Equation~\ref{eq:3} will encounter the weight misalignment problem, which is overlooked by existing methods. 

To investigate potential conflicts when merging a specific layer, we measure its contribution by calculating the performance difference before and after the merge. Precisely, we assess the merging contribution from two directions:

\begin{itemize}[leftmargin=*]
    \item \textbf{Deletion Impact $(\alpha)$}: To estimate this impact, we first construct a merged model $\mathcal{M}_G$ that merges all layers using the merging process mentioned in Equation~\eqref{eq:1} or Equation~\eqref{eq:3}. Then, we calculate the performance degradation caused by removing the delta vector for a specific layer.
    \item \textbf{Addition Impact ($\beta$)}: This impact is measured by the performance improvement of adding the delta vector for a specific layer to the pre-trained foundation model $\mathcal{M}_F$.
\end{itemize}

These impacts can be mathematically represented as:
\begin{align}
    \alpha_{m1,m2}^{l} = &\mathrm{P}_{t_{m1}}(\boldsymbol{\theta}_{m2}- \boldsymbol{\hat\delta}_{m2}^l) - \mathrm{P}_{t_{m1}}(\boldsymbol{\theta}_{m2}) \label{eq:ConAna_start},\\
    \beta_{m1,m2}^{l} = &\mathrm{P}_{t_{m1}}(\boldsymbol{\hat\theta}_{F} + \boldsymbol{\delta}_{m2}^l) - \mathrm{P}_{t_{m1}}(\boldsymbol{\theta}_{F}),
\end{align}
where $m1\in\{A,B\}$ is the task capability identifier and $m_2\in\{A,B,G\}$ is the model identifier. We investigate the layer-wise contribution so that $l$ is the layer index. $\boldsymbol{\hat\delta}_{m2}^l$ is the delta vector for $\mathcal{M}_{m2}$ at layer $l$. $\mathrm{P}_{t_{m1}}(\cdot)$ represents the performance metric on the task ${t_{m1}}$. For example, BLEU-4~\cite{bleu} score for the QA task.

We sum up two impacts as the overall contribution:
\begin{equation}
    c_{m1,m2}^{l} = \alpha_{m1,m2}^{l}+ \beta_{m1,m2}^{l}\label{eq:ConAna_end}.
\end{equation}



\subsubsection{Iterative Conflict Elimination}

\begin{figure}[t]
    \centering
    \includegraphics[width=\columnwidth]{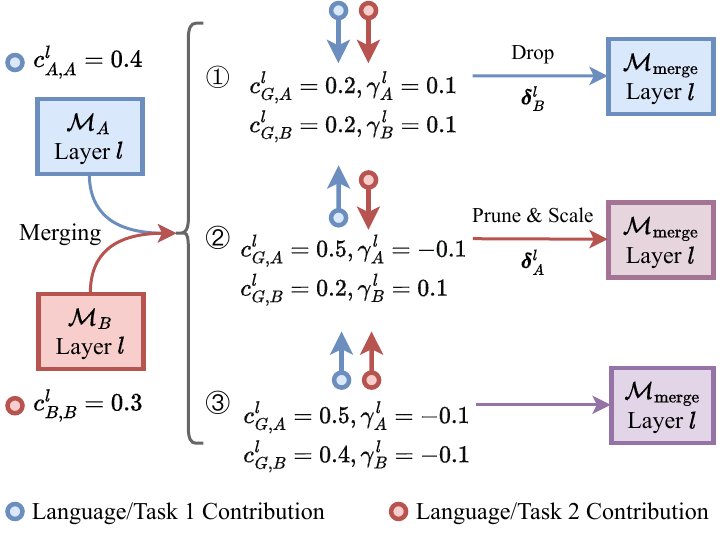}
    \caption{The demonstration of different conflict elimination strategies for three pre-merging conditions.}
    \label{fig:strategy}
\end{figure}
The contribution analysis method defined in Equation~\eqref{eq:ConAna_start}-\eqref{eq:ConAna_end} provides a solution to measure the importance of merging specific layers. We can then define the conflict resulted by model $\mathcal{M}_m (m\in \{A,B\})$ within the layer $l$ of the merged model as:
\begin{align}
    \gamma^l_{m} =c_{m,m}^{l} - c_{m,G}^{l}.
\end{align}
In this formula, we set the capability identifier $m1=m$ as we expect the merged model can maintain the performance of $\mathcal{M}_m$ on $t_m$. 
We can then identify the most severe conflicting layers that impair the fine-tuned performance by sorting $\Gamma^l =\sum_{m\in \{A,B\}} \gamma^l_{m}$. 

To mitigate the parameter misalignment, we iteratively merge the most conflicting layers (with the largest $\Gamma^l$).
Specifically, to process a specific layer, there are three types of conflict as illustrated in Figure~\ref{fig:strategy}:
\begin{enumerate}[leftmargin=*]
    \item \textbf{Severe Conflict:} $\gamma^l_{A}>0$ and $\gamma^l_{B}>0$, indicating both capabilities are impaired by the merging. In such cases, only the delta vector with a larger contribution is retained, e.g., dropping $\boldsymbol{\hat\delta}_{B}^l$ in the figure. Namely,  $\boldsymbol{\hat\delta}_{B}^l$ is set to zero.
    \item \textbf{Partial Conflict:} $\gamma^l_{A}*\gamma^l_{B}<0$, i.e., one of the delta vectors leads to the parameter misalignment. The solution for this case is to prune and scale the conflict delta vector again, as we defined in Section~\ref{model-wise-pruning-scaling}. For example, in Figure~\ref{fig:strategy}, the overfitting on $t_A$ ($\gamma^l_{A}<0$ and $\gamma^l_{B}>0$) leads to the degradation of the ability for $t_B$. As a result, we prune and scale $\boldsymbol{\hat \delta}^l_{A}$ again as\footnote{We use the same notation $\boldsymbol{\hat \delta}^l_{A}$ for clarity.}:
    \begin{equation}
        \boldsymbol{\hat\delta}^l_{A} = s_{A} \cdot \operatorname{Top}_{p_{A}}(\boldsymbol{\hat \delta}^l_{A}).
    \end{equation}
    \item \textbf{Mutual Enhancement:} 
    If $\gamma^l_{A}\le0$ and $\gamma^l_{B}\le0$, the merging process improves for both capabilities. In this case, no further adjustment is necessary for this layer.
\end{enumerate}
After resolving the conflicts of all layers, the parameters of the final merged model $\mathcal{M}_{merge}$ is:
\begin{equation}
\boldsymbol{\theta}_{\mathrm{merge}} = \boldsymbol{\theta}_{\mathrm{F}} + \boldsymbol{\hat\delta}_{A} + \boldsymbol{\hat\delta}_{B}.
\end{equation}


\section{Experiments}
\label{experiments}

In this section, we conduct comprehensive experiments to evaluate the effectiveness of \mname by answering following research questions (RQ):
\begin{itemize}[leftmargin=*]
    \item \textbf{RQ1:} How does \mname perform when merging LLMs that excel at the same task but in different languages?
    \item \textbf{RQ2:} How does \mname perform when merging LLMs excel at different tasks with the same languages?
    \item \textbf{RQ3:} Is \mname applicable for merging LLMs across languages and tasks?
    \item \textbf{RQ4:} Can \mname effectively merge different open-source LLMs?
    \item \textbf{RQ5:} How is the merging conflict under our method’s settings?
    \item \textbf{RQ6:} What is the impact of different components of \mname on its overall performance?
\end{itemize}

\subsection{Experimental Settings}

\subsubsection{Datasets}
We select four datasets listed in Table~\ref{tab:datasets} that cover multilingual multi-task capabilities, including English and Chinese languages, with multiple-choice question answering (MCQA) and open-domain question answering (QA) tasks.

\subsubsection{Baselines}

In our experiments, we use the multilingual and multi-task models fine-tuned on combined datasets as strong baselines. For model merging approaches, we consider a range of general model merging methods, including weighted averaging (Model Soups~\cite{modelsoup}) and delta vector-based approaches (Arithmetic~\cite{taskarithmetic}, TIES-Merging~\cite{ties}, DARE~\cite{dare}, DELLA~\cite{della}, and Model Breadcrumbs~\cite{modelbread}). We further compare with two knowledge transfer approaches: OT-Fusion~\cite{otfusion} and Layer Swapping~\cite{layerswap}.
Details are in Appendix~\ref{sssec:appendix-baselines}.

\begin{table}[t]
    \centering
    \caption{The brief description and statistics of the four datasets (MedQA~\cite{medqa}, CMExam~\cite{cmexam}, HealthCareMagic~\cite{chatdoctor}, and cMedQA2)~\cite{cMedQA2} used for fine-tuning.}
    \label{tab:datasets}

    \resizebox{\columnwidth}{!}{
    \setlength{\tabcolsep}{3pt}
    \begin{tabular}{lccccc}
        \toprule
        \textbf{Name} & \textbf{Task} & \textbf{Language} & \textbf{Train} & \textbf{Validation} & \textbf{Test} \\
        \midrule
        MedQA                & MCQA & English  & 10,000   & 400   & 400    \\
        CMExam               & MCQA & Chinese  & 50,000  & 4,000 & 4,000  \\
        HealthCareMagic      & QA   & English  & 30,000 & 1,000 & 1,000  \\
        cMedQA2              & QA   & Chinese  & 30,000 & 1,000 & 1,000  \\
        \bottomrule
    \end{tabular}
    }
\end{table}

\subsubsection{Implementation Details}
\label{ssec:implementation-details}
We use Qwen2-7B-Instruct as foundation models
with results for other foundation models presented in Appendix~\ref{sec:appendix-multilingual}. For fine-tuning, we employ LLaMA-Factory~\footnote{\href{https://github.com/hiyouga/LLaMA-Factory}{https://github.com/hiyouga/LLaMA-Factory}} with LoRA (rank=8, alpha=16, dropout=0.01) and a batch size of 64. The learning rate is $1.0^{-4}$ with cosine decay and warm-up. LLM merging is performed using  mergekit~\footnote{\href{https://github.com/arcee-ai/mergekit}{https://github.com/arcee-ai/mergekit}}. Both $p$ and $s$ in model-wise process range from 0.1 to 1.0 with a step of 0.1. In layer-wise process, the pruning threshold $p$ and scaling factor $s$ are successively set to half of their model-wise values.


\subsubsection{Evaluation Metrics}

For the MCQA task, accuracy is employed to measure the proportion of correct answers~\cite{bert}.
For the QA task, we use BLEU-4~\cite{bleu} to evaluate the precision of the generation, and ROUGE-1,2,L~\cite{rouge} to assess the overlap and coherence with the ground truth. 
Additionally, we report both the average relative performance improvement (Avg Impr.) and the mean ranking (Avg Rank.) across all methods.

\subsection{Bilingual Task Merging (RQ1)}

\begin{table}[t]
    \centering
    \caption{
    Performance comparison of merging methods for bilingual MCQA task. Model A is fine-tuned on MedQA. Model B is fine-tuned on CMExam. Multi-task model is fine-tuned on both. The overall best result is in bold and the best merging result is underlined.
    }
    \label{tab:multilingual-mcqa}
    \resizebox{\linewidth}{!}{
    \begin{tabular}{@{}llcccc@{}}
        \toprule 
        \textbf{Types} & \textbf{Methods} & \textbf{\makecell{L1\\ (MedQA)}} & \textbf{\makecell{L2\\ (CMExam)}} & \textbf{Avg Impr.} & \textbf{Avg Rank.} \\ \midrule
        \multirow{3}{*}{Pre-trained} & Qwen2-7B-Instruct & 51.4062 & 74.6217 & -        & 17.0 \\
                                     & Yi-1.5-9B         & 46.8185 & 58.6499 & -16.31\% & 18.0 \\
                                     & Baichuan2-7B      & 6.4415  & 7.1439  & -89.22\% & 19.0 \\ \midrule
        \multirow{3}{*}{Fine-tuned}  & Model A (L1)      & 59.1406 & 83.7771 & +13.40\% & 10.0 \\
                                     & Model B (L2)      & 54.4531 & 88.6171 & +13.52\% & 11.5 \\
                                     & Multi-task        & 60.0781 & 88.2246 & +17.67\% & 3.5  \\ \midrule
        \multirow{10}{*}{Merged}     & Model Soups       & 59.6094 & 88.6926 & +17.67\% & 5.0  \\
                                     & Task Arithmetic   & 59.5312 & 88.7681 & +17.67\% & 4.0  \\
                                     & TIES              & 59.0625 & 88.7832 & +17.31\% & 4.5  \\
                                     & DARE              & 58.6719 & 88.6926 & +16.93\% & 7.5  \\
                                     & DARE + TIES       & 58.9063 & 88.6021 & +17.04\% & 8.0  \\
                                     & Model Breadcrumbs & 58.8281 & 88.6322 & +17.00\% & 8.5  \\
                                     & DELLA             & 58.9844 & 88.7681 & +17.24\% & 5.5  \\
                                     & DELLA + TIES      & 58.2812 & 88.7530 & +16.67\% & 9.5  \\
                                     & OT-Fusion         & 59.8271 & 88.6543 & +17.60\% & 3.0  \\
                                     & Layer Swapping    & 55.6406 & 87.0859 & +16.24\% & 16.0 \\
                                     & \mname (Ours)     & {\ul \textbf{60.1562}} & {\ul \textbf{89.0700}} & {\ul \textbf{+18.41\%}} & {\ul \textbf{1.0}} \\ \bottomrule
    \end{tabular}
    }
\end{table}

We first verify the effectiveness of \mname on bilingual task merging. Here, we merge models trained on the MCQA task in English and Chinese, as shown in Table~\ref{tab:multilingual-mcqa}. Additional experiments on the QA task and a different LLM are provided in Appendix~\ref{sec:appendix-multilingual} due to space constraints.

Baseline methods like Model Soups and Task Arithmetic that combine models without considering noises and conflicts achieve stable but lower performance. Methods that reduce conflicts, such as TIES and DARE, occasionally achieve the best results on individual metrics. However, without a clear guidance, their performance highly randomised. In contrast, our \mname method, with hierarchical pruning and scaling approach, not only achieves the best average performance but attains optimal results in about half of the individual metrics.
We also investigate the impact of different training sample sizes in Appendix~\ref{ssec:appendix-samples}.

\subsection{Monolingual Multi-task Merging (RQ2)}

\begin{table*}[ht]
\centering
\caption{
Performance comparison of merging methods for tasks with different question formats. Model A is fine-tuned on MCQA tasks (T1), while model B is fine-tuned on QA tasks (T2). The overall best result is marked in bold and the best merging result is underlined.
}
\label{tab:multi-task}
\resizebox{\textwidth}{!}{
\begin{tabular}{@{}l|l|ccccc|ccccc|c|c@{}}
\toprule
\multirow{3}{*}{\textbf{Types}} & \multirow{3}{*}{\textbf{Methods}} & \multicolumn{5}{c|}{\textbf{L1 (English)}} & \multicolumn{5}{c|}{\textbf{L2 (Chinese)}} & \multirow{3}{*}{\textbf{\makecell{Avg \\ Impr.} }} & \multirow{3}{*}{\textbf{\makecell{Avg \\ Rank.}}} \\ \cmidrule(lr){3-12}
 &  & \multicolumn{1}{c|}{\textbf{T1 (MedQA)}} & \multicolumn{4}{c|}{\textbf{T2 (HealthCareMagic)}} & \multicolumn{1}{c|}{\textbf{T1 (CMExam)}} & \multicolumn{4}{c|}{\textbf{T2 (cMedQA2)}} &  &  \\ \cmidrule(lr){3-12}
 &  & \multicolumn{1}{c|}{Accuracy} & BLEU-4 & ROUGE-1 & ROUGE-2 & ROUGE-l & \multicolumn{1}{c|}{Accuracy} & BLEU-4 & ROUGE-1 & ROUGE-2 & ROUGE-l &  &  \\ \midrule
Pre-trained & Qwen2-7B-Instruct & \multicolumn{1}{c|}{51.4062} & 30.1209 & 26.3524 & 5.3280 & 15.7451 & \multicolumn{1}{c|}{74.6217} & 1.7090 & 14.1527 & 1.7822 & 9.0934 & - & - \\ \midrule
Fine-tuned & Model A (T1) & \multicolumn{1}{c|}{59.1406} & 34.6533 & 28.7482 & 6.9168 & 17.9525 & \multicolumn{1}{c|}{88.6171} & 2.8064 & 16.8617 & 2.5603 & 12.0561 & +17.36\% & 11.2 \\
 & Model B (T2) & \multicolumn{1}{c|}{53.0469} & 35.5717 & 30.2512 & 8.9044 & 20.3625 & \multicolumn{1}{c|}{81.5670} & 4.4159 & 21.2210 & 4.0680 & 17.4600 & +20.21\% & 7.2 \\
 & Multi-task & \multicolumn{1}{c|}{59.2188} & 35.6009 & 30.2101 & 9.1375 & 20.4645 & \multicolumn{1}{c|}{88.6926} & 3.7790 & 20.5919 & 3.8096 & 16.9265 & +25.23\% & 8.3 \\ \midrule
Merged & Model Soups & \multicolumn{1}{c|}{58.5156} & 36.4411 & 30.5654 & 9.1754 & 20.4259 & \multicolumn{1}{c|}{88.8285} & 4.3912 & 21.0216 & 4.0040 & 17.2843 & +26.19\% & 5.6 \\
 & Task Arithmetic & \multicolumn{1}{c|}{58.5938} & 36.3290 & {\ul \textbf{30.6624}} & {\ul \textbf{9.1945}} & {\ul \textbf{20.5406}} & \multicolumn{1}{c|}{88.7983} & 4.3018 & 20.6467 & 3.7496 & 16.9995 & +26.13\% & 6.1 \\
 & TIES & \multicolumn{1}{c|}{60.4688} & 35.7851 & 30.3243 & 9.0310 & 20.3723 & \multicolumn{1}{c|}{88.6171} & 4.5434 & {\ul \textbf{21.5629}} & 4.1910 & 17.4909 & +26.78\% & 4.2 \\
 & DARE & \multicolumn{1}{c|}{58.4375} & {\ul \textbf{36.5802}} & 30.5488 & 9.0818 & 20.3945 & \multicolumn{1}{c|}{88.7681} & 4.5487 & 21.3255 & 3.8403 & 17.4471 & +26.29\% & 4.4 \\
 & DARE+TIES & \multicolumn{1}{c|}{59.3750} & 35.7062 & 30.1950 & 8.7840 & 20.0878 & \multicolumn{1}{c|}{88.8285} & 4.1587 & 21.1291 & 3.8124 & 17.2868 & +25.63\% & 7.5 \\
 & Model Breadcrumbs & \multicolumn{1}{c|}{57.8906} & 36.4620 & 30.2173 & 8.7845 & 20.0169 & \multicolumn{1}{c|}{88.8889} & 4.4472 & 21.2492 & 3.8931 & 17.2846 & +25.53\% & 6.4 \\
 & DELLA & \multicolumn{1}{c|}{58.5938} & 36.3494 & 30.1715 & 8.8125 & 20.1879 & \multicolumn{1}{c|}{88.8134} & 4.3718 & 21.0226 & 3.9300 & 17.3403 & +25.83\% & 7.1 \\
 & DELLA+TIES & \multicolumn{1}{c|}{59.5312} & 36.0774 & 30.4743 & 9.1151 & 20.4599 & \multicolumn{1}{c|}{88.6021} & 4.3202 & 21.2269 & 4.0164 & 17.3779 & +25.96\% & 5.7 \\
 & \mname (Ours) & \multicolumn{1}{c|}{{\ul \textbf{60.5469}}} & 36.4926 & 30.5467 & 9.1231 & 20.3523 & \multicolumn{1}{c|}{{\ul \textbf{88.9795}}} & {\ul \textbf{4.6781}} & 21.5367 & {\ul \textbf{4.2165}} & {\ul \textbf{17.5038}} & {\ul \textbf{+27.07\%}} & {\ul \textbf{2.1}} \\ \bottomrule
\end{tabular}
}
\end{table*}

\begin{table*}[ht]
\centering
\caption{Performance comparison of merging methods for bilingual multi-task learning. Model A is fine-tuned on MCQA datasets (T1: MedQA or CMExam). Model B is fine-tuned on QA datasets (T2: cMedQA2 or HealthCareMagic). The overall best result is marked in bold and the best merging result is underlined.}
\label{tab:multilingual-multi-task}
\resizebox{\textwidth}{!}{
\begin{tabular}{@{}l|l|c|cccc|c|cccc|c|c@{}}
\toprule
\multirow{2}{*}{\textbf{Types}} & \multirow{2}{*}{\textbf{Methods}} & \textbf{\begin{tabular}[c]{@{}c@{}}T1, L1\\ (MedQA)\end{tabular}} & \multicolumn{4}{c|}{\textbf{\begin{tabular}[c]{@{}c@{}}T2, L2\\ (cMedQA2)\end{tabular}}} & \textbf{\begin{tabular}[c]{@{}c@{}}T1, L2\\ (CMExam)\end{tabular}} & \multicolumn{4}{c|}{\textbf{\begin{tabular}[c]{@{}c@{}}T2, L1\\ (HealthCareMagic)\end{tabular}}} & \multirow{2}{*}{\textbf{\makecell{Avg \\ Impr.}}} & \multirow{2}{*}{\textbf{\makecell{Avg \\ Rank.}}} \\ \cmidrule(lr){3-12}
 &  & Accuracy & BLEU-4 & ROUGE-1 & ROUGE-2 & ROUGE-l & Accuracy & BLEU-4 & ROUGE-1 & ROUGE-2 & ROUGE-l &  &  \\ \midrule
Pre-trained & Qwen2-7B-Instruct & 51.4062 & 1.7090 & 14.1527 & 1.7822 & 9.0934 & 74.6217 & 30.1209 & 26.3524 & 5.3280 & 15.7451 & - & - \\ \midrule
Fine-tuned & Model A (T1) & 59.1406 & 2.8064 & 16.8617 & 2.5603 & 12.0561 & 88.6171 & 34.6713 & 28.4279 & 6.6122 & 18.1117 & +17.17\% & 10.7 \\
 & Model B (T2) & 54.4922 & 4.4159 & 21.2210 & 4.0680 & 17.4600 & 79.6875 & 35.5717 & 30.2512 & 8.9044 & 20.3625 & +20.03\% & 7.2 \\
 & Multi-task & \textbf{60.7812} & 3.8473 & 20.8741 & 4.0434 & 16.9525 & \textbf{88.9795} & 35.7429 & 30.1735 & 8.9153 & 20.3902 & +26.22\% & 6.8 \\ \midrule
Merged & Model Soups & 58.3584 & 4.6592 & 21.2316 & 4.0559 & 17.3805 & 88.6322 & 36.1765 & {\ul \textbf{30.7169}} & {\ul \textbf{9.2702}} & {\ul \textbf{20.5227}} & +26.35\% & 4.5 \\
 & Task Arithmetic & 58.0469 & 4.6682 & 21.2618 & 4.0984 & 17.4231 & 88.7379 & 36.1222 & 30.2256 & 8.7570 & 20.1357 & +25.69\% & 5.7 \\
 & TIES & 59.6094 & 4.3764 & 21.0083 & 3.9002 & 17.4194 & 88.7228 & 35.7708 & 30.5143 & 8.8994 & 20.3487 & +26.16\% & 6.4 \\
 & DARE & 57.8906 & 4.5671 & 21.1856 & 3.9549 & 17.2328 & 88.6322 & 35.8639 & 30.1489 & 8.8150 & 20.1025 & +25.22\% & 8.1 \\
 & DARE+TIES & 58.75 & 4.4929 & 21.3194 & 4.0824 & 17.4826 & 88.5568 & 34.8223 & 29.7597 & 8.3004 & 19.7624 & +24.76\% & 7.8 \\
 & Model Breadcrumbs & 57.1094 & 4.7217 & 21.4192 & 4.1477 & 17.4182 & 88.6021 & 36.4961 & 30.3911 & 9.0696 & 20.4108 & +25.82\% & 4.3 \\
 & DELLA & 58.0469 & {\ul \textbf{4.8065}} & {\ul \textbf{21.5135}} & 4.1356 & 17.4962 & 88.6167 & 36.0159 & 30.3747 & 9.0414 & 20.3929 & +26.11\% & {\ul \textbf{3.9}} \\
 & DELLA+TIES & 59.0625 & 4.4854 & 20.9954 & 4.0491 & {\ul \textbf{17.5630}} & 88.6624 & 35.0176 & 29.9666 & 8.6580 & 20.1406 & +25.31\% & 7.5 \\
 & \mname (Ours) & {\ul 60.2344} & 4.7743 & 21.1954 & {\ul \textbf{4.1749}} & 17.3991 & {\ul 88.7983} & {\ul \textbf{36.5223}} & 30.3932 & 8.7882 & 20.1619 & {\ul \textbf{+27.02\%}} & 4.1 \\ \bottomrule
\end{tabular}
}
\end{table*}

\begin{figure}[t]
    \centering
    \includegraphics[width=\columnwidth]{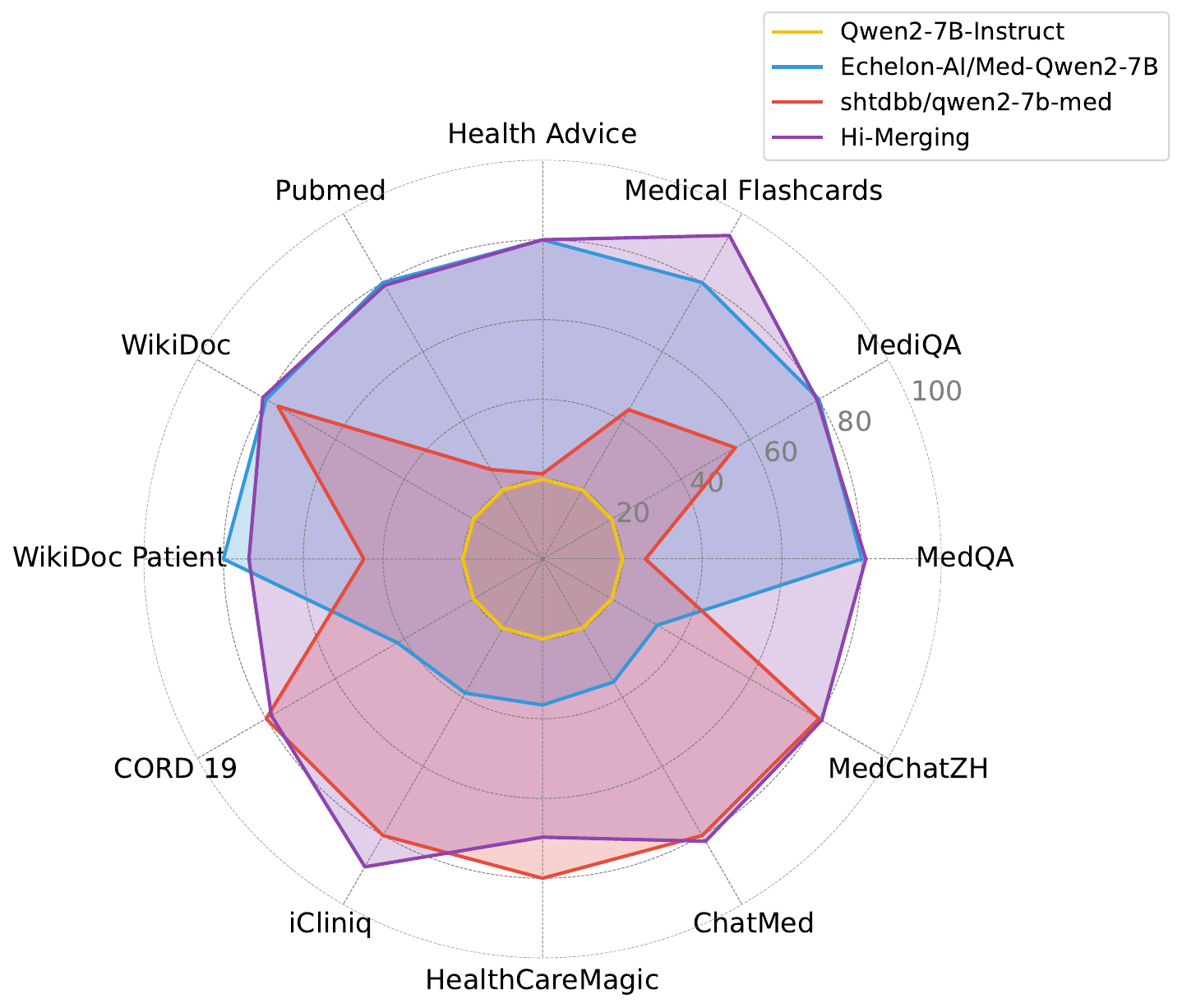}
    \caption{Performance of Hi-Merging on two open-source medical models, Echelon-AI/Med-Qwen2-7B and shtdbb/qwen2-7b-med, which are fine-tuned from the foundation model Qwen/Qwen2-7B-Instruct.}
    \label{fig:general}
\end{figure}

For monolingual multi-task merging, we combine models trained on different tasks with the same language (e.g., English MCQA with English QA), as shown in Table~\ref{tab:multi-task}.
The results show that merged models consistently outperform their individual fine-tuned counterparts, with many even surpassing multi-task fine-tuned models. Notably, our \mname approach achieves a 1.84\% relative improvement over the multi-task fine-tuned model. 
We attribute this success to three factors. 1) During multi-task fine-tuning with limited data (compared to pre-training), tasks can interfere with each other due to the ``seesaw effect''. In contrast, model merging allows parameters to be optimized independently before integration, avoiding such interference. 2) Since both models are fine-tuned from the same foundation model, their parameter updates tend to follow similar optimization trajectories, making successful merging more likely. 3) The inherent sparsity of LLMs provides sufficient parameter space to accommodate multi-task knowledge from both models.

\subsection{Bilingual Multi-task Merging (RQ3)}

For bilingual multi-task merging, we combine models trained on completely different tasks and languages. Specifically, we merge a model trained for MCQA in one language (Model A: MedQA in English or CMExam in Chinese) with another model trained for QA in the opposite language (Model B: cMedQA2 in Chinese or HealthCareMagic in English), as illustrated in Table~\ref{tab:multilingual-multi-task}.

Our experiments reveal an interesting pattern: bilingual multi-task fine-tuning mainly affects QA performance, while MCQA performance remain. This can be explained by two factors: (1) QA tasks require complex free-form generation, making them more vulnerable to joint fine-tuning; (2) MCQA tasks involve clear classification boundaries and simpler choice selection, making them more robust to merging process.

\begin{table}[t]
    \centering
    \caption{
    Performance of \mname on two open-source medical models: Echelon-AI/Med-Qwen2-7B and shtdbb/qwen2-7b-med.
    }
    \label{tab:merge-medical}
    \resizebox{\linewidth}{!}{
        \begin{tabular}{@{}llccc@{}}
        \toprule
        \multirow{2}{*}{\textbf{Types}} & \multirow{2}{*}{\textbf{Models}} & \multicolumn{2}{c}{\textbf{Medical}} & \textbf{Math} \\ \cmidrule(l){3-5} 
         &  & \textbf{MedQA} & \textbf{Pubmed} & \textbf{GSM8K} \\ \midrule
        Single & Pre-trained & 37.3868 & 5.7994 & \textbf{65.96} \\
         & Echelon-AI/Med-Qwen2-7B & 64.2862 & \textbf{92.9898} & 15.92 \\
         & shtdbb/qwen2-7b-med & 40.1598 & 11.5017 & 57.77 \\
        Merged & \mname (Medical) & \textbf{64.9011} & 92.1692 & 56.63 \\ \bottomrule
        \end{tabular}
    }
\end{table}

\begin{table}[t]
    \centering
    \caption{
    Performance of \mname after merging the medical model with the math-specialized model Qwen2-Math-7B-Instruct.
    }
    \label{tab:merge-med-math}
    \resizebox{\linewidth}{!}{
        \begin{tabular}{@{}llccc@{}}
        \toprule
        \multirow{2}{*}{\textbf{Types}} & \multirow{2}{*}{\textbf{Models}} & \multicolumn{2}{c}{\textbf{Medical}} & \textbf{Math} \\ \cmidrule(l){3-5} 
         &  & \textbf{MedQA} & \textbf{Pubmed} & \textbf{GSM8K} \\ \midrule
        Single & Pre-trained & 37.3868 & 5.7994 & 65.96 \\
        Merged & Merged (Medical) & \textbf{64.9011} & \textbf{92.1692} & 56.63 \\
         & Qwen2-Math-7B-Instruct & 36.7716 & 15.3618 & \textbf{79.00} \\
        Merged & \mname (Medical + Math) & 63.6742 & 91.7320 & 77.45 \\ \bottomrule
        \end{tabular}
    }
\end{table}

\subsection{Open-source LLM Merging (RQ4)}

To validate the generality of our merging approach, we conduct experiments using two open-source medical models from Hugging Face: 
Echelon-AI/Med-Qwen2-7B~\footnote{\href{https://huggingface.co/Echelon-AI/Med-Qwen2-7B}{https://huggingface.co/Echelon-AI/Med-Qwen2-7B}}, fine-tuned on English datasets for tasks such as medical QA and information retrieval (IR), and shtdbb/qwen2-7b-med~\footnote{\href{https://huggingface.co/shtdbb/qwen2-7b-med}{https://huggingface.co/shtdbb/qwen2-7b-med}}, fine-tuned on Chinese datasets for dialogue generation. Both models are derived from Qwen2-7B-Instruct.
Figure~\ref{fig:general} illustrates the performance comparison across 12 medical datasets, with metrics normalized for better visualization. 

Our approach demonstrates robust performance across the task spectrum. In 7 out of 12 datasets, \mname achieves the best performance among all models, with only two datasets showing apparent degradation compared to the better-performing individual model.
These results demonstrate \mname's ability to effectively fuse medical knowledge while maintaining or enhancing performance across diverse languages and tasks. Detailed implementation setup and unprocessed numerical results can be found in Appendix~\ref{sssec:appendix-implementation-details} and \ref{ssec:appendix-open-source}. 

To further demonstrate the adaptability of our approach in other domains, we merge the combined medical model with an additional mathematical model Qwen/Qwen2-Math-7B-Instruct~\footnote{\href{https://huggingface.co/Qwen/Qwen2-Math-7B-Instruct}{https://huggingface.co/Qwen/Qwen2-Math-7B-Instruct}}. As shown in Table~\ref{tab:merge-medical} and Table~\ref{tab:merge-med-math}, merging two models from the same domain (medical) leads to mutually beneficial integration. However, when further merging the medical model with an LLM from a different domain (math), we observe slight performance drops on both domains, suggesting that larger divergence in fine-tuning data increases the difficulty of effective knowledge integration.

\subsection{Case Study (RQ5)}

\begin{figure}[t]
    \centering
    \includegraphics[width=\columnwidth]{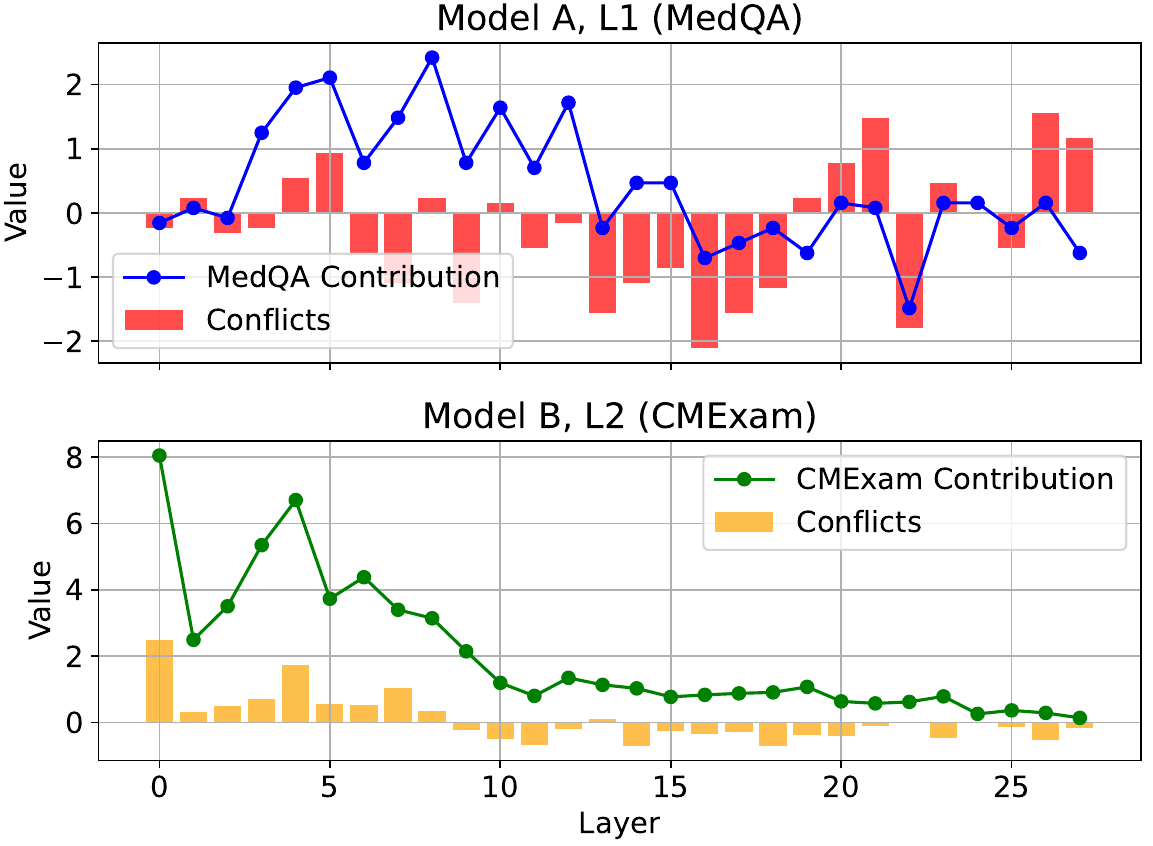}
    \caption{The visualization of layer contributions and merging conflicts when merging model fine-tuned on MedQA and CMExam.}
    \label{fig:case}
\end{figure}

Figure~\ref{fig:case} visualizes layer-wise contributions and merging conflicts when combining MedQA and CMExam models, revealing that conflicts are not uniformly distributed. Model A (MedQA) shows significant conflicts in later layers, while Model B (CMExam) exhibits conflicts in earlier layers. This non-uniformity highlights the need for \mname's hierarchical pruning and scaling strategy, leading to the improved performance demonstrated in previous experiments in Table~\ref{tab:multilingual-qa}. Such layer-specific conflict patterns suggest that different layers may specialize in different tasks, making a uniform merging strategy suboptimal.

\subsection{Ablation Study (RQ6)}

\begin{table}[ht]
    \centering
    \caption{
    Ablation study of different processes in \mname for bilingual MCQA task merging. Model A is fine-tuned on MedQA. Model B is fine-tuned on CMExam. Multi-task model is fine-tuned on both. The overall best result is in bold.
    }
    \label{tab:ablation}
    \resizebox{\linewidth}{!}{
    \begin{tabular}{@{}llccc@{}}
        \toprule
        \textbf{Types} & \textbf{Methods} & \textbf{\makecell{L1\\ (MedQA)}} & \textbf{\makecell{L2\\ (CMExam)}} & \textbf{Avg Impr.}  \\  \midrule
        Pre-trained & Qwen2-7B-Instruct & 51.4062 & 74.6217 & - \\ \midrule
        \multirow{3}{*}{Fine-tuned} & Model A (L1) & 59.1406 & 83.7771 & +13.40\% \\
         & Model B (L2) & 54.4531 & 88.6171 & +13.52\% \\
         & Multi-task & 60.0781 & 88.2246 & +17.67\% \\ \midrule
        Merged & w/o All & 59.5312 & 88.5291 & +17.48\% \\
         & w/o Model-wise Process & 59.8437 & 88.6501 & +17.83\% \\
         & \quad   w/o Model-wise Pruning & 60.0781 & 88.9342 & +18.24\% \\
         & \quad w/o Model-wise Scaling & 59.9219 & 88.7863 & +18.00\% \\
         & w/o Layer-wise Process & 59.6094 & 88.5417 & +17.55\% \\
         & \quad w/o Layer-wise Pruning & \textbf{61.0156} & 88.6473 & \textbf{+18.75\%} \\
         & \quad w/o Layer-wise Scaling & 59.7656 & 88.6926 & +17.80\% \\
         & \mname & 60.1562 & \textbf{89.0700} & +18.41\% \\ \bottomrule
    \end{tabular}
    }
\end{table}

The ablation study in Table~\ref{tab:ablation} reveals several key insights. Layer-wise process has a more significant impact than model-wise process, and removing scaling operations leads to larger performance drops than removing pruning. While removing layer-wise pruning achieves the highest average improvement, it shows less consistent performance across tasks compared to the full \mname approach, indicating that pruning helps stabilize the merging process despite potentially limiting peak performance on specific tasks.

\section{Related Works}
\label{related-works}

\subsection{Multilingual Task-Oriented LLMs}

Multi-task learning (MTL) has proven valuable across various domains, from recommendation systems~\cite{single-shot, DBLP:conf/sigir/Zhang0Y00ZLLZH024} to knowledge graphs~\cite{DBLP:conf/coling/XuZLWZ000C24} and healthcare~\cite{DBLP:journals/corr/abs-2402-02803}, by enabling models to share knowledge between related tasks. This paradigm becomes especially relevant for multilingual NLP, where different languages face similar challenges in tasks like machine translation~\cite{machinetranslation}, text summarization~\cite{textsummarization}, and sentiment analysis~\cite{sentimentanalysis}. 

Recently, LLMs have greatly contributed to advancing multilingual tasks by leveraging massive amounts of multilingual data \cite{gpt3,bert,mt5}. Despite their success, LLMs exhibit a clear performance gap across languages: they excel at widely-spoken languages with abundant training data but struggle with less-represented languages that have limited online presence~\cite{mtl-survey}.

To enhance multilingual capabilities, LLMs employ continual training on specific languages, as seen in models like Chinese-LLaMA~\cite{chinese-llama-alpaca} and EuroLLM~\cite{EuroLLM}. Additionally, supervised fine-tuning techniques, such as LoRA in Chinese-Alpaca~\cite{chinese-llama-alpaca}, further improve multilingual understanding. However, LLMs are usually enhanced for one language at a time, resulting in multiple isolated models.

\subsection{Model Merging}

Model merging aims to integrate knowledge from multiple fine-tuned models into a single one. These methods are categorized into two types: weighted-based merging and interference mitigation.

Weighted-based merging focuses on combining model parameters effectively. This includes simple techniques like parameter averaging, such as Model Soups~\cite{modelsoup}, Fisher-weighted merging~\cite{fisher} and RegMean~\cite{regmean}. While computationally efficient, these methods often miss conflicting parameter updates, leading to performance degradation. Therefore, Task Arithmetic \cite{taskarithmetic} proposes manipulating delta vectors. AdaMerging \cite{AdaMerging} and evolutionary algorithms \cite{evolution} optimize merging coefficients and blend diverse models, respectively.

Interference mitigation techniques aim to reduce parameter conflicts based on the over-parameterization and sparsity of LLM. SparseGPT \cite{SparseGPT} show high LLM performance despite significant parameter pruning. DELLA \cite{della} introduces MAGPRUNE for selective pruning and parameter rescaling. 
However, these techniques focus mainly on individual parameter-level operations without considering the structural relationships and knowledge dependencies across model layers.

\section{Conclusion}
\label{sec:conclusion}

In this paper, we proposed \mname, a novel approach for merging LLMs for multilingual multi-task learning. \mname leverages model-wise and layer-wise pruning and scaling strategy to minimize the conflict between fine-tuned models' delta vectors.
The model-wise process eliminates the fine-tuning noise and overfitting parameters of the original models.
Then, the layer-wise process analyzes the contribution of each layer’s delta vector to the fine-tuning performance, reducing the interference of conflicts in several key layers.
Extensive experiments on the MCQA and QA datasets demonstrated that \mname outperforms traditional merging techniques and even surpasses models trained on multiple datasets.  
Future work will explore finer-grained conflict analysis strategies.

\section{Limitations}

While our proposed Hi-Merging method demonstrates promising results, several limitations should be acknowledged. First, our current method only supports merging two models at a time. Extending the approach to simultaneously merge multiple models presents additional challenges in terms of conflict resolution and computational complexity, which requires further investigation.

Second, our evaluation is currently limited to two task types (MCQA and QA) and two languages (English and Chinese). The effectiveness of Hi-Merging on a broader range of NLP tasks and language families remains to be investigated. This includes exploring its applicability to tasks such as text generation, summarization, and semantic parsing across diverse language groups.

Third, our method focuses on merging models fine-tuned from the same foundation model. The applicability and performance of Hi-Merging when merging models from different architectural families or pre-training approaches is yet to be explored. This limitation becomes particularly relevant as the field continues to see diverse model architectures and training paradigms.

Finally, our current implementation assumes relatively balanced task importance. The method might need adaptation for scenarios where certain tasks or languages should be prioritized over others, potentially requiring a more flexible weighting mechanism in the merging process. Future work could explore dynamic weighting strategies that adapt to specific application requirements and performance objectives.

\section*{Acknowledgments}

This research was partially supported by Research Impact Fund (No.R1015-23), Collaborative Research Fund (No.C1043-24GF) and Tencent (CCF-Tencent Open Fund, Tencent Rhino-Bird Focused Research Program).

\bibliography{custom}

\appendix

\section{Appendix}
\label{sec:appendix}

\subsection{Experimental Settings}

\subsubsection{Baselines}
\label{sssec:appendix-baselines}

In our experiments, we compare it against a comprehensive set of baseline methods, including traditional weighted averaging techniques and state-of-the-art approaches specifically developed for fine-tuned models.

\begin{itemize}[leftmargin=*]

\item \textbf{Multilingual Multi-task Training} This approach trains a single model on the combined datasets of multiple languages simultaneously, without distinguishing between tasks.

\item \textbf{Model Soups}~\cite{modelsoup} Uniform Soup is a simple merging method where the parameters of the fine-tuned models are averaged based on their importance.

\item \textbf{Task Arithmetic}~\cite{taskarithmetic} This method performs arithmetic operations on the parameter differences between the pre-trained and fine-tuned models.

\item \textbf{TIES}~\cite{ties} The Task Interference Elimination Strategy (TIES) minimize negative transfer and task interference by pruning redundant parameters and using a chosen sign to determine parameter update directions.

\item \textbf{DARE}~\cite{dare} Delta Alignment for Robust Ensemble (DARE) reduces the interference across tasks by randomly drop the delta vectors.

\item \textbf{Model Breadcrumbs}~\cite{modelbread} This approach tracks and prunes maxima and minima in delta vectors to retain critical task-specific features.

\item \textbf{DELLA}~\cite{della} DELLA follows DARE and assign drop rates to delta vectors according to their absolute values, improving performance stability.

\item \textbf{OT-Fusion}~\cite{otfusion} This method aligns and averages model weights via optimal transport, enabling one-shot parameter merging across heterogeneous models without retraining.

\item \textbf{Layer Swapping}~\cite{layerswap} This approach composes task and language experts by directly replacing top and bottom transformer layers, facilitating cross-lingual transfer.

\end{itemize}

\begin{table*}[t]
    \centering
    \caption{
    Performance comparison of merging methods for bilingual QA tasks. Model A is fine-tuned on HealthCareMagic, Model B is fine-tuned on cMedQA2, Multi-task model is fine-tuned on both datasets. The overall best result is marked in bold and the best merging result is underlined.
    }
    \label{tab:multilingual-qa}
    \resizebox{\textwidth}{!}{
    \begin{tabular}{@{}l|l|cccc|cccc|c|c@{}}
    \toprule
    \multirow{2}{*}{\textbf{Types}} & \multirow{2}{*}{\textbf{Methods}} & \multicolumn{4}{c|}{\textbf{L1 (HealthCareMagic)}} & \multicolumn{4}{c|}{\textbf{L2 (cMedQA2)}} & \multirow{2}{*}{\textbf{Avg Impr.}} & \multirow{2}{*}{\textbf{Avg Rank.}} \\ \cmidrule(lr){3-10}
     &  & BLEU-4 & ROUGE-1 & ROUGE-2 & ROUGE-l & BLEU-4 & ROUGE-1 & ROUGE-2 & ROUGE-l &  &  \\ \midrule
    Pre-trained & Qwen2-7B-Instruct & 30.1209 & 26.3524 & 5.3280 & 15.7451 & 1.7090 & 14.1527 & 1.7822 & 9.0934 & - & - \\ \midrule
    \multirow{3}{*}{Fine-tuned} & Model A (L1) & 35.5717 & \textbf{30.2512} & \textbf{8.9044} & 20.3625 & 3.7609 & 19.1370 & 3.1364 & 15.1441 & +30.66\% & 6.875 \\
     & Model B (L2) & 24.8587 & 24.9841 & 4.1492 & 15.1967 & 4.4159 & 21.2210 & 4.0680 & 17.4600 & +11.57\% & 8.375 \\
     & Multi-task & 35.7637 & 29.9781 & 8.6687 & 20.1184 & 3.7660 & 20.9869 & 3.7784 & 16.8850 & +34.19\% & 6.125 \\ \midrule
    \multirow{9}{*}{Merged} & Model Soups & 33.2627 & 28.8258 & 7.5487 & 18.9459 & 4.6801 & {\ul \textbf{21.5564}} & 4.0502 & 17.5380 & +30.80\% & 6.125 \\
     & Task Arithmetic & 33.0398 & 28.7169 & 7.5726 & 18.9600 & 4.7181 & 21.4108 & 4.0503 & 17.6772 & +30.55\% & 5.625 \\
     & TIES & 33.6571 & 29.0496 & 7.7769 & 19.1503 & 4.3751 & 20.8551 & 3.7518 & 17.1978 & +30.23\% & 7.375 \\
     & DARE & 33.3031 & 28.9575 & 7.8222 & 19.1702 & {\ul \textbf{4.7578}} & 21.0865 & 3.8996 & 17.2488 & +30.64\% & 5.625 \\
     & DARE+TIES & 26.8091 & 26.0330 & 5.2307 & 16.5201 & 4.2456 & 20.6276 & 3.7531 & 17.1445 & +15.41\% & 10.375 \\
     & Model Breadcrumbs & 34.3247 & 29.4403 & 8.1518 & 19.6443 & 4.4092 & 20.9365 & 3.8138 & 17.1378 & +32.19\% & 6.750 \\
     & DELLA & 33.4207 & 28.9234 & 7.6728 & 18.9674 & 4.6827 & 21.1596 & 4.0709 & 17.4775 & +30.77\% & 5.500 \\
     & DELLA+TIES & 27.2331 & 26.1723 & 5.4339 & 16.6009 & 4.7130 & 21.2275 & {\ul \textbf{4.2944}} & {\ul \textbf{17.7694}} & +18.37\% & 6.000 \\
     & \mname (Ours) & {\ul \textbf{35.9500}} & {\ul 29.9826} & {\ul 8.8738} & {\ul \textbf{20.3844}} & 4.7009 & 21.1752 & 3.9704 & 17.2361 & {\ul \textbf{+36.42\%}} & {\ul \textbf{3.250}} \\ \bottomrule
    \end{tabular}
    }
\end{table*}

\begin{table*}[t]
    \centering
    \caption{
    Performance comparison of merging methods for multilingual QA using Llama-3-8B-Instruct. Model A is fine-tuned on HealthCareMagic (L1: English). Model B is fine-tuned on cMedQA2 (L2: Chinese). Multi-task model is fine-tuned on both datasets. The overall best result is marked in bold and the best merging result is underlined.
    }
    \label{tab:multilingual-qa-llama3}
    \resizebox{\textwidth}{!}{
    \begin{tabular}{@{}l|l|cccccccc|c|c@{}}
    \toprule
    \multirow{2}{*}{\textbf{Types}} & \multirow{2}{*}{\textbf{Methods}} & \multicolumn{4}{c}{\textbf{L1 (HealthCareMagic)}} & \multicolumn{4}{c|}{\textbf{L2 (cMedQA2)}} & \multirow{2}{*}{\textbf{Avg.}} & \multirow{2}{*}{\textbf{Impr.}} \\ \cmidrule(lr){3-10}
     &  & BLEU-4 & ROUGE-1 & ROUGE-2 & \multicolumn{1}{c|}{ROUGE-L} & BLEU-4 & ROUGE-1 & ROUGE-2 & ROUGE-L &  &  \\ \midrule
    Pre-trained & Llama-3-8B-Instruct & 16.3118 & 21.6011 & 3.1389 & \multicolumn{1}{c|}{10.8666} & 0.0225 & 0.4710 & 0.0211 & 0.2343 & 6.5834 & - \\ \midrule
    \multirow{3}{*}{Fine-tuned} & Model A (L1) & \textbf{36.0325} & 30.4111 & \textbf{9.2743} & \multicolumn{1}{c|}{\textbf{20.7236}} & 0.0185 & 0.1288 & 0.0025 & 0.0841 & 12.0844 & +83.5\% \\
     & Model B (L2) & 3.9950 & 7.7673 & 0.9267 & \multicolumn{1}{c|}{4.6358} & 3.0638 & 20.4016 & 3.4178 & 16.3067 & 7.5643 & +14.9\% \\
     & Multi-task & 35.6154 & \textbf{30.5447} & 9.2156 & \multicolumn{1}{c|}{20.4271} & 3.0250 & 20.3136 & \textbf{3.4964} & 16.0911 & \textbf{17.3411} & \textbf{+163.5\%} \\ \midrule
    \multirow{9}{*}{Merged} & Model Soups & 32.1199 & 28.2278 & 6.3715 & \multicolumn{1}{c|}{18.2456} & 3.3256 & 19.5499 & 2.8670 & 15.4688 & 15.7720 & +139.5\% \\
     & Task Arithmetic & 31.6679 & 27.7646 & 6.0354 & \multicolumn{1}{c|}{18.0448} & 3.3805 & 19.6475 & 2.9507 & 15.4806 & 15.6215 & +137.2\% \\
     & TIES & 32.1494 & 28.0527 & 6.7440 & \multicolumn{1}{c|}{18.2913} & 3.2238 & 19.5369 & 2.8854 & 15.2112 & 15.7618 & +139.4\% \\
     & DARE & 25.9679 & 25.6716 & 4.5173 & \multicolumn{1}{c|}{16.3803} & 3.5337 & 20.8586 & 3.1736 & 16.6716 & 14.5968 & +121.7\% \\
     & DARE+TIES & 26.6707 & 25.9106 & 5.2031 & \multicolumn{1}{c|}{16.5525} & 3.2236 & 19.8564 & 2.9963 & 15.5967 & 14.5012 & +120.2\% \\
     & Model Breadcrumbs & 26.9844 & 26.1004 & 4.7037 & \multicolumn{1}{c|}{16.3247} & 3.3307 & 20.7442 & 3.3069 & 16.2874 & 14.7228 & +123.6\% \\
     & DELLA & 25.6313 & 25.6792 & 4.5522 & \multicolumn{1}{c|}{16.1313} & {\ul \textbf{3.6612}} & {\ul \textbf{20.9176}} & {\ul 3.3286} & {\ul \textbf{16.7355}} & 14.5796 & +121.4\% \\
     & DELLA+TIES & 27.1246 & 26.0186 & 5.3163 & \multicolumn{1}{c|}{16.6170} & 3.3433 & 19.9122 & 3.0848 & 15.9942 & 14.6764 & +122.9\% \\
     & \mname (Ours) & {\ul 33.5960} & {\ul 28.4141} & {\ul 7.2167} & \multicolumn{1}{c|}{{\ul 18.8804}} & 3.1967 & 19.8207 & 2.9509 & 15.7833 & {\ul 16.2324} & {\ul +146.5\%} \\ \bottomrule
    \end{tabular}
    }
\end{table*}

\begin{table*}[ht]
\caption{Numerical performance of Hi-Merging on two open-source models, Echelon-AI/Med-Qwen2-7B and shtdbb/qwen2-7b-med.}
\label{tab:appendix-open}
\resizebox{\textwidth}{!}{
\begin{tabular}{@{}lcccccccccccc@{}}
\toprule
\textbf{Models} & \textbf{MedQA} & \textbf{MediQA} & \textbf{Medical Flashcards} & \textbf{Health Advice} & \textbf{Pubmed} & \textbf{WikiDoc} & \textbf{WikiDoc Patient} & \textbf{CORD 19} & \textbf{iCliniq} & \textbf{HealthCareMagic} & \textbf{ChatMed} & \textbf{MedChatZH} \\ \midrule
Qwen2-7B-Instruct & 37.3868 & 17.3595 & 22.7668 & 2.8205 & 5.7994 & 17.6217 & 18.785 & 39.1748 & 19.3292 & 28.7051 & 9.9138 & 8.0654 \\
Echelon-AI/Med-Qwen2-7B & 64.2862 & \textbf{32.052} & 41.1081 & 97.7523 & \textbf{92.9898} & 20.7237 & \textbf{26.9203} & 40.7167 & 26.5593 & 30.3212 & 15.1218 & 9.2714 \\
shtdbb/qwen2-7b-med & 40.1598 & 27.1442 & 29.85 & 4.096 & 11.5017 & 20.5808 & 21.1528 & \textbf{41.2026} & 27.332 & \textbf{33.3678} & 19.4513 & 11.2665 \\
\mname (Ours) & \textbf{64.9011} & 31.9421 & \textbf{45.1714} & \textbf{97.755} & 92.1692 & \textbf{21.0211} & 26.3293 & 40.9803 & \textbf{28.7816} & 31.6779 & \textbf{19.8074} & \textbf{11.2958} \\ \bottomrule
\end{tabular}
}
\end{table*}

\begin{table}[t]
    \centering
    \caption{
    Performance comparison of merging methods for bilingual MCQA using Llama-3-8B-Instruct. Model A is fine-tuned on MedQA (L1: English). Model B is fine-tuned on CMExam (L2: Chinese), Multi-task model is fine-tuned on both datasets. Overall best result is in bold and the best merging result is underlined.
    }
    \label{tab:multilingual-mcqa-llama3}
    \resizebox{\linewidth}{!}{
    \begin{tabular}{@{}llcccc@{}}
        \toprule
        \textbf{Types} & \textbf{Methods} & \textbf{\makecell{L1\\ (MedQA)}} & \textbf{\makecell{L2\\ (CMExam)}} & \textbf{Avg.} & \textbf{Impr.} \\ \midrule
        \multirow{3}{*}{Pre-trained} 
            & Llama-3-8B-Instruct & 57.9733 & 17.2821 & 37.6277 & +0.00\% \\
            & GLM-4-9B            & 54.7656 & 69.5194 & 62.1425 & \textbf{+65.15\%} \\
            & Gemma-2-9B          & 14.2583 & 2.7698  & 8.5141  & -77.37\% \\ \midrule
        \multirow{3}{*}{Fine-tuned} 
            & Model A (L1)        & 60.4688 & 52.2706 & 56.3697 & +49.81\% \\
            & Model B (L2)        & 60.3906 & 60.5525 & 61.0575 & +62.27\% \\
            & Multi-task        & \textbf{62.8906} & 61.0356 & \textbf{61.9631} & +64.70\% \\ \midrule
        \multirow{9}{*}{Merged} 
            & Model Soups         & 61.2500 & 61.0507 & 61.1504 & +62.54\% \\
            & Task Arithmetic     & 61.2500 & {\ul \textbf{61.8750}} & 61.5625 & +63.65\% \\
            & TIES                & 61.7188 & 61.3225 & 61.5207 & +63.56\% \\
            & DARE                & 61.5625 & 61.3678 & 61.4652 & +63.42\% \\
            & DARE + TIES         & 60.9375 & 59.4656 & 60.2016 & +60.05\% \\
            & Model Breadcrumbs   & 61.0156 & 60.4318 & 60.7237 & +61.43\% \\
            & DELLA               & 60.8594 & 60.7186 & 60.7890 & +61.58\% \\
            & DELLA + TIES        & 61.9531 & 61.3527 & 61.6529 & +63.91\% \\
            & \mname (Ours)       & {\ul 62.2656} & 61.0757 & {\ul 61.6707} & {\ul +63.96\%} \\ 
        \bottomrule
    \end{tabular}
    }
\end{table}

\subsubsection{Implementation Details}
\label{sssec:appendix-implementation-details}


For model adaptation, we applied LoRA to all linear networks in the model. The learning rate schedule was carefully designed with a 100-step warm-up phase followed by cosine decay, which helped achieve stable convergence while maintaining optimal model performance. This configuration proved effective in balancing training efficiency and model quality across both multilingual and multi-task scenarios.

In addition to Qwen2-7B-Instruct, we also experimented with other foundation models including Llama-3-8B-Instruct (results shown in \ref{sec:appendix-multilingual}). However, Qwen2-7B-Instruct demonstrated more consistent performance, particularly in handling both English and Chinese tasks, making it the preferred choice for our main experiments.

For visualization in Figure~\ref{fig:general}, we normalized the performance metrics to facilitate clear comparisons. The performance values of the models on each dataset represent the average of the QA task metrics (BLEU-4, ROUGE-1, ROUGE-2, and ROUGE-L). We scaled the pre-trained Qwen2-7B-Instruct's performance to 20 and the better-performing fine-tuned model's performance to 80 for each task. The performance values of the other fine-tuned model and our merged model were then proportionally adjusted within this range to maintain their relative differences.

\begin{figure*}[t]
    \centering
    \begin{subfigure}[b]{0.245\textwidth}
        \includegraphics[width=\textwidth]{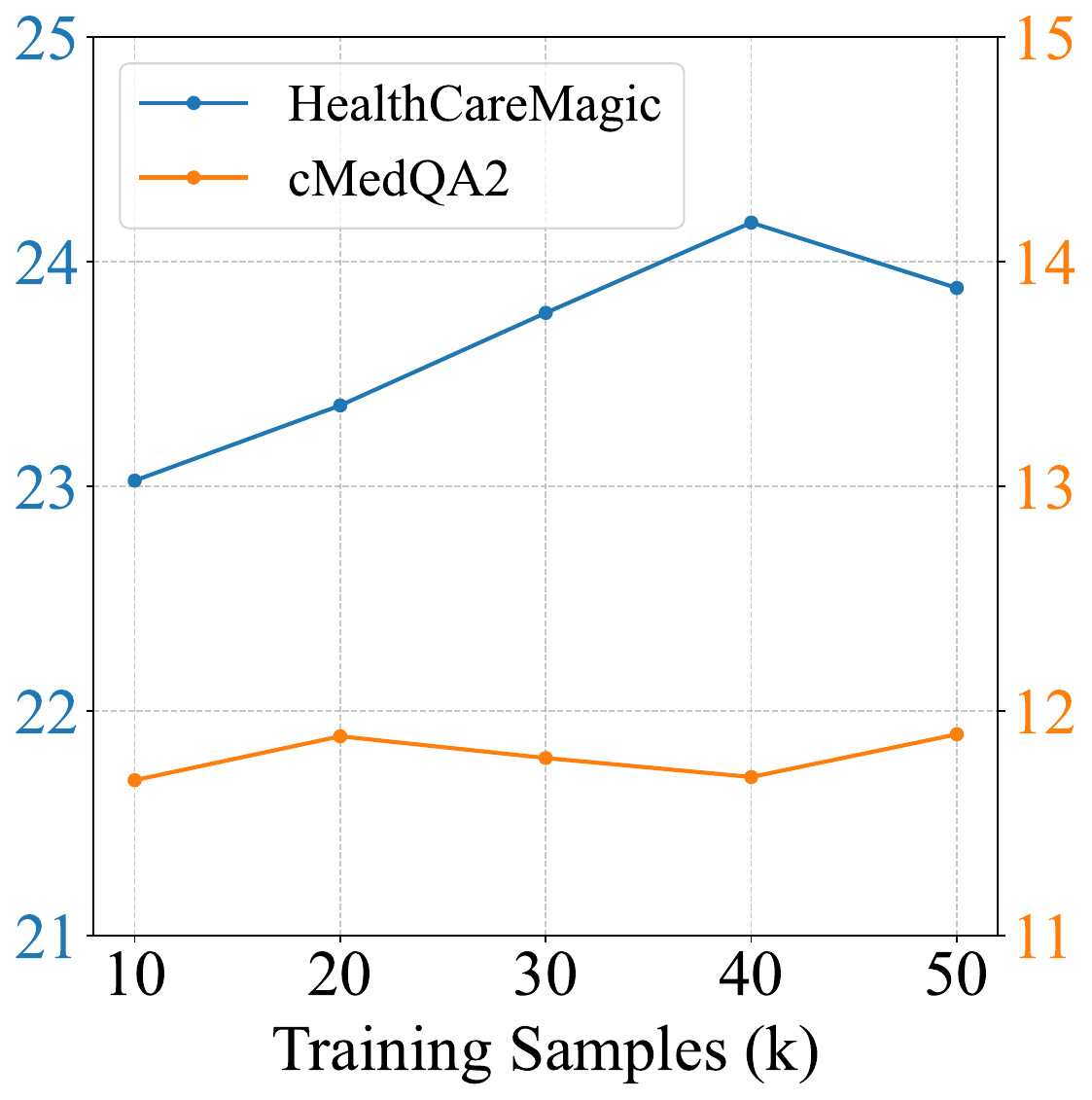}
        \caption{Fine-tuned models}
        \label{fig:sample1}
    \end{subfigure}
    \hfill
    \begin{subfigure}[b]{0.245\textwidth}
        \includegraphics[width=\textwidth]{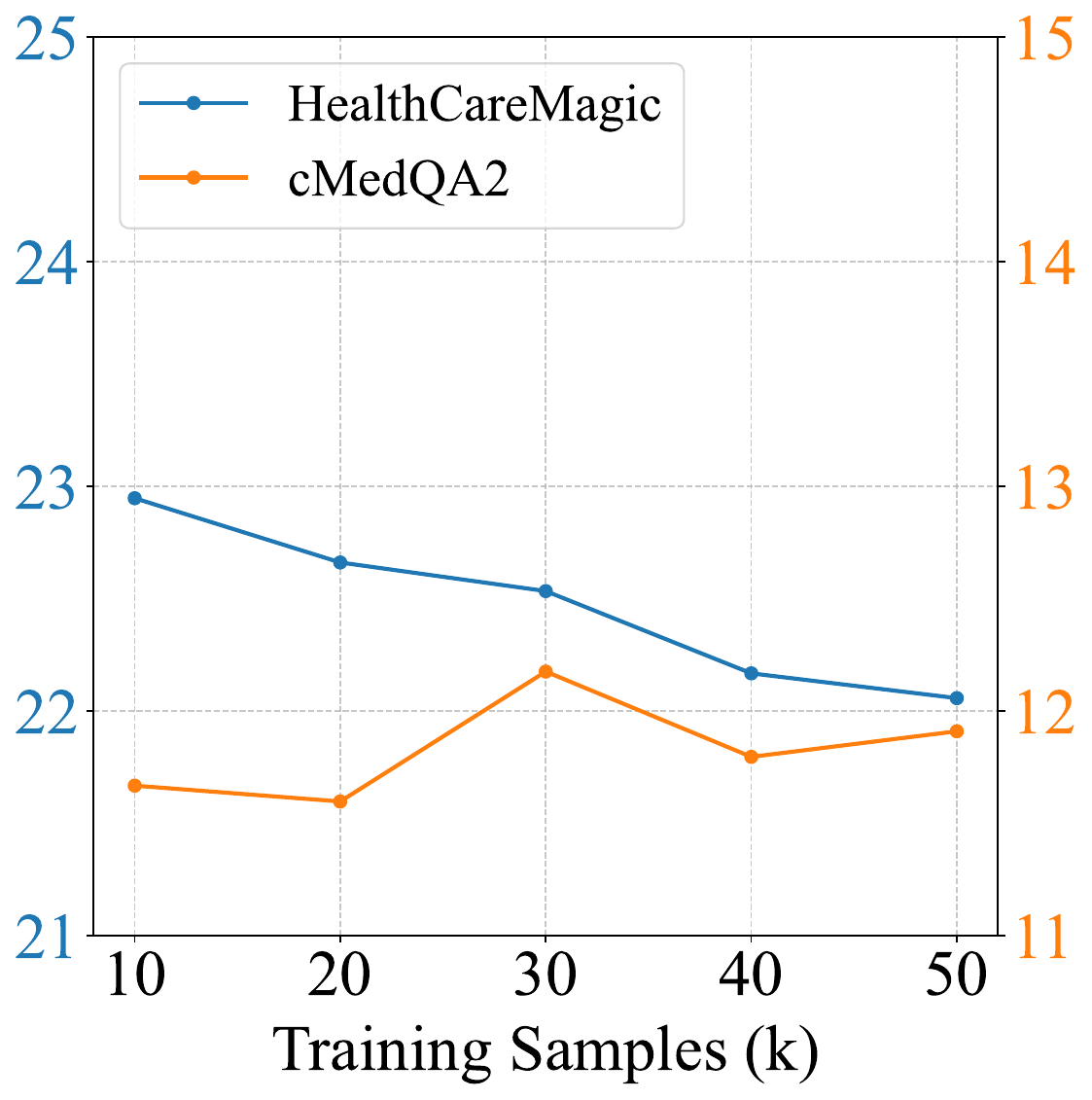}
        \caption{Model Soups}
        \label{fig:sample2}
    \end{subfigure}
    \hfill
    \begin{subfigure}[b]{0.245\textwidth}
        \includegraphics[width=\textwidth]{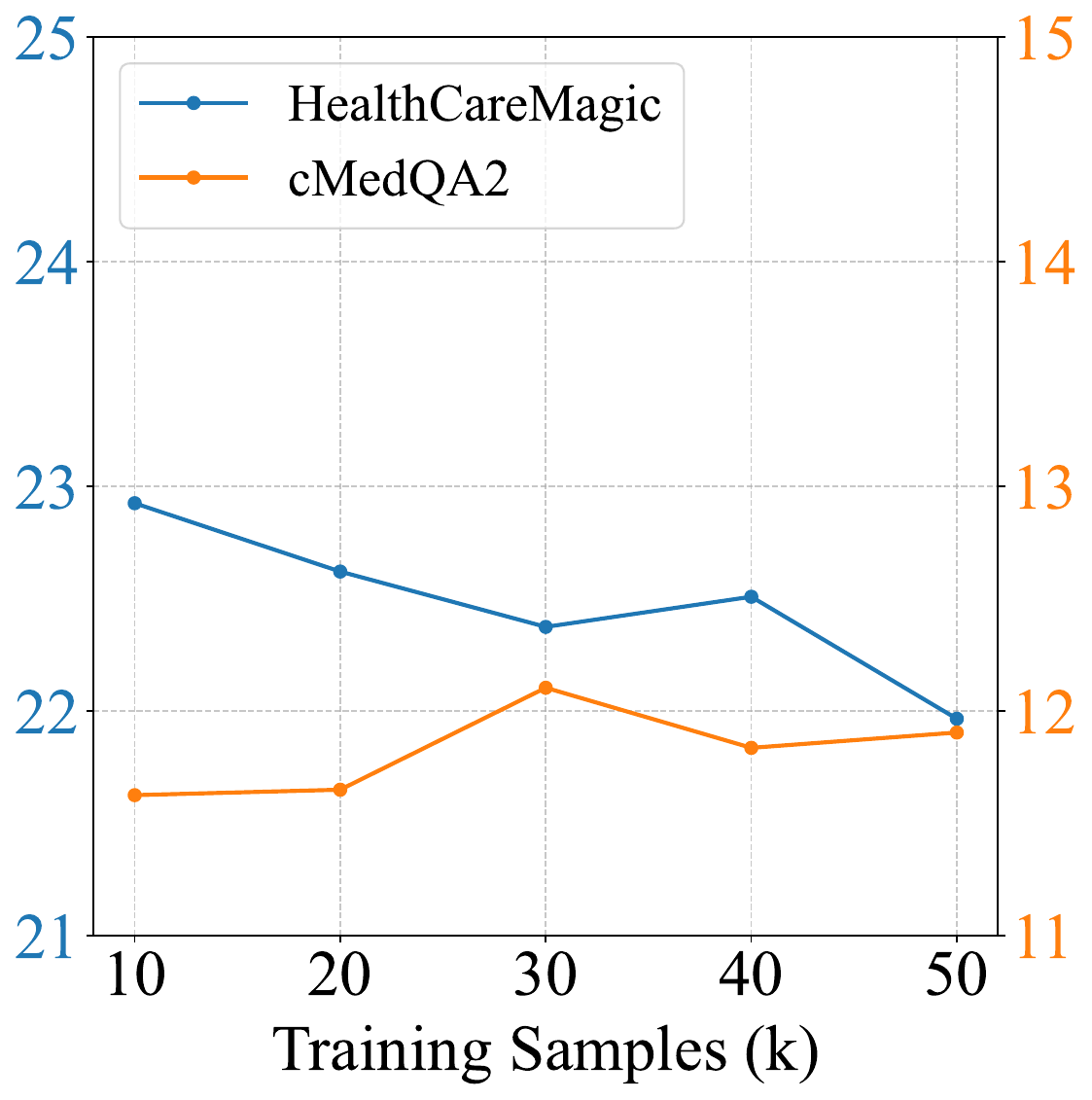}
        \caption{Task Arithmetic}
        \label{fig:sample3}
    \end{subfigure}
    \hfill
    \begin{subfigure}[b]{0.245\textwidth}
        \includegraphics[width=\textwidth]{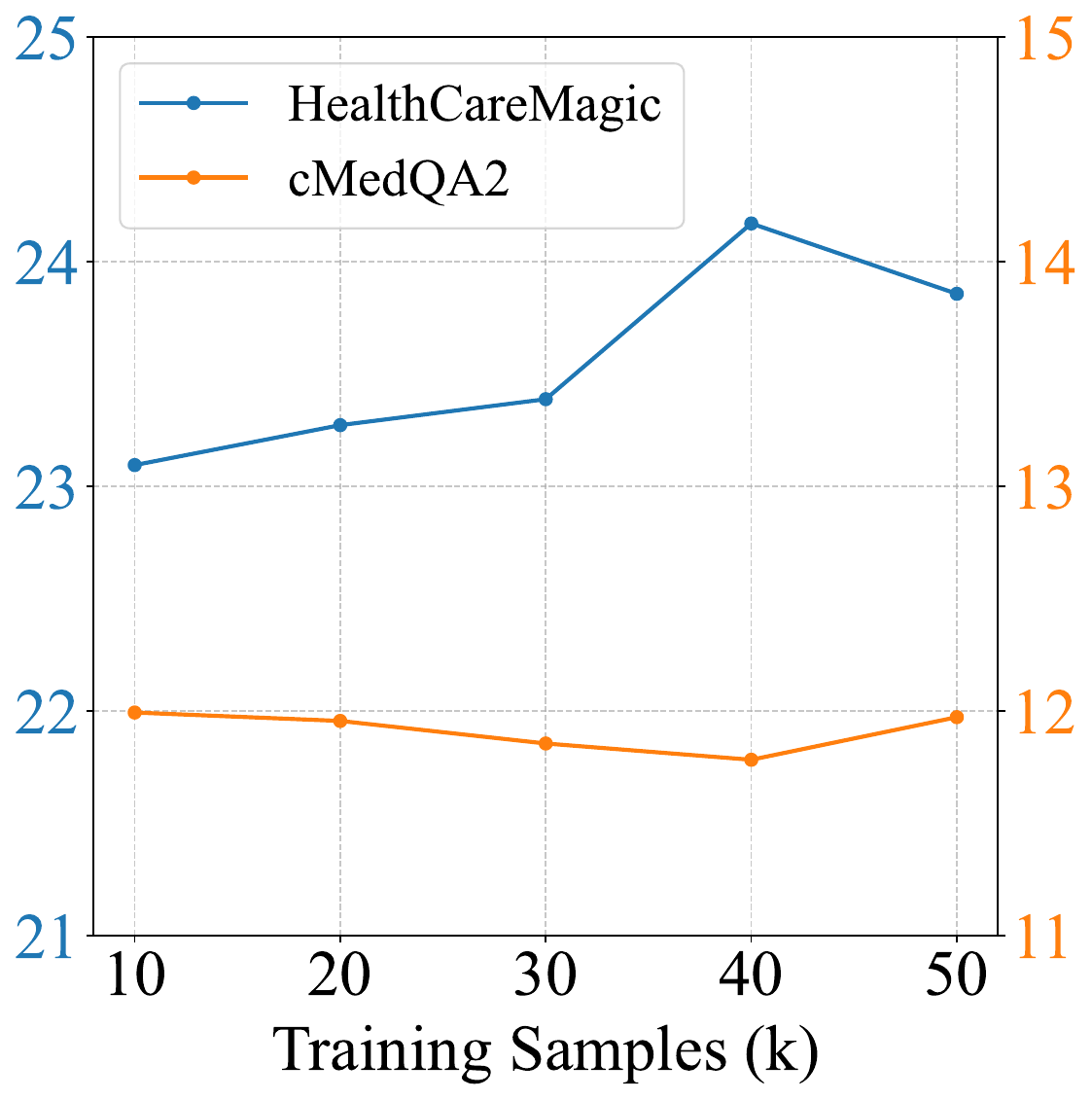} 
        \caption{Our method}
        \label{fig:sample4}
    \end{subfigure}
    \caption{Impact of training sample size on model merging conflicts. Blue and orange lines represent the average performance metrics for HealthCareMagic and cMedQA2, respectively.}
    \label{fig:training_samples}
\end{figure*}

\subsection{Bilingual Task Merging}
\label{sec:appendix-multilingual}

For merging LLMs that specialise in different languages on the same task, we further conduct experiments on the QA task (Table~\ref{tab:multilingual-qa}) and extend the foundation LLM to Llama-3-8B-Instruct, as presented in Table~\ref{tab:multilingual-mcqa-llama3} and \ref{tab:multilingual-qa-llama3}. 

The results in Table~\ref{tab:multilingual-qa} demonstrate the effectiveness of our approach in merging bilingual QA models. \mname achieves the best performance on English QA metrics (BLEU-4: 35.95, ROUGE-1: 29.98, ROUGE-2: 8.87, ROUGE-L: 20.38) while maintaining competitive performance on Chinese QA metrics. This balanced performance leads to the highest average improvement (+36.42\%) and best average ranking (3.25) among all merging methods. Notably, while some baseline methods like DELLA+TIES achieve better performance on specific Chinese metrics, they significantly compromise English performance, highlighting our method's advantage in maintaining cross-lingual capabilities.

The results in Table~\ref{tab:multilingual-qa-llama3} show that the performance of merged models based on Llama-3-8B-Instruct is generally inferior to that of the pre-merged fine-tuned models. This indicates that the effectiveness of the merging process is strongly influenced by the quality of the foundation models.
The observed degradation in performance can be attributed to several factors. First, weaker foundation models, such as Llama-3-8B-Instruct, tend to produce delta vectors with more dispersed and less coherent parameter distributions during fine-tuning. These delta vectors often carry noisy or conflicting information, which makes the merging process prone to parameter conflicts. Second, the weaker representational capacity of these models limits their ability to encode robust and semantically aligned knowledge, further exacerbating the challenges of merging.


\subsection{Number of training samples}
\label{ssec:appendix-samples}

We examine the impact of varying the number of training samples on the conflict during model merging, as shown in Figure~\ref{fig:training_samples}.
In the experiment, we use two QA datasets, HealthCareMagic (English) and cMedQA2 (Chinese), sampling 10k, 20k, 30k, 40k, and 50k training examples from each to produce a series of fine-tuned models, five per dataset. 
This setup evaluates how the number of training samples influences both individual model performance and compatibility during merging. 
The x-axis of Figure~\ref{fig:training_samples} represents the number of training samples, while the y-axis denotes the average performance metrics, including BLEU-4, ROUGE-1, ROUGE-2, and ROUGE-L.

However, Figures~\ref{fig:sample2} and \ref{fig:sample3} show that merged models through either Model Soups or Task Arithmetic suffer from performance drops driven by the increasing size of training sample as further training leads to conflicting highly specialized models. Figure~\ref{fig:sample4} shows the opposite: our method retains performance trends in line with fine-tuned models and addresses conflicts to retain improving performance through larger training sets.

These results highlight the robustness of our method in resolving merging conflicts, ensuring that the merged models retain the strengths of individual models while achieving stable and superior performance across training sample sizes.

\subsection{Open Source LLM Merging}
\label{ssec:appendix-open-source}

Table~\ref{tab:appendix-open} presents the detailed numerical results for all models across the 12 medical datasets. The datasets cover a wide range of medical tasks and languages, allowing us to comprehensively evaluate the models' capabilities and the effectiveness of our merging approach.










\end{document}